\documentclass{article}

\usepackage[preprint]{neurips_2024_style}
\setcitestyle{numbers}

\usepackage{microtype}
\usepackage{graphicx}
\usepackage{subfigure}
\usepackage{caption}
\usepackage{subcaption}
\usepackage{float}
\usepackage{booktabs} 

\usepackage[utf8]{inputenc} 
\usepackage[T1]{fontenc}    
\usepackage{hyperref}       
\usepackage{url}            
\usepackage{booktabs}       
\usepackage{amsfonts}       
\usepackage{nicefrac}       
\usepackage{microtype}      
\usepackage{xcolor}         
\usepackage{amsmath}
\usepackage{amsthm}
\usepackage{amssymb}
\usepackage{mathrsfs}
\usepackage{MnSymbol}

\newtheorem{definition}{Definition}

\usepackage{multirow}
\newtheorem{theorem}{Theorem}

\usepackage{enumitem}

\title{Teleporter Theory: A General and Simple Approach for Modeling Cross-World Counterfactual Causality}

\author{%
  Jiangmeng Li\thanks{Contributed equally to this work.}, Bin Qin$^*$, Qirui Ji, Yi Li, Wenwen Qiang\thanks{Corresponding author.}, Jianwen Cao, Fanjiang Xu \\
  Institute of Software\\
  Chinese Academy of Sciences\\
  Beijing, China \\
  \texttt{qiangwenwen@iscas.ac.cn} \\
}

\begin{document}

\maketitle

\begin{abstract}
Leveraging the development of structural causal model (SCM), researchers can establish graphical models for exploring the causal mechanisms behind machine learning techniques. As the complexity of machine learning applications rises, \textit{single-world} interventionism causal analysis encounters theoretical adaptation limitations. Accordingly, \textit{cross-world} counterfactual approach extends our understanding of causality beyond observed data, enabling hypothetical reasoning about alternative scenarios. However, the joint involvement of cross-world variables, encompassing counterfactual variables and real-world variables, challenges the construction of the graphical model. Twin network is a subtle attempt, establishing a symbiotic relationship, to bridge the gap between graphical modeling and the introduction of counterfactuals albeit with room for improvement in generalization. In this regard, we demonstrate the theoretical \textit{breakdowns} of twin networks in certain cross-world counterfactual scenarios. To this end, we propose a novel \textit{teleporter theory} to establish a general and simple graphical representation of counterfactuals, which provides criteria for determining \textit{teleporter} variables to connect multiple worlds. In theoretical application, we determine that introducing the proposed teleporter theory can directly obtain the conditional independence between counterfactual variables and real-world variables from the cross-world SCM without requiring complex algebraic derivations. Accordingly, we can further identify counterfactual causal effects through cross-world symbolic derivation. We demonstrate the generality of the teleporter theory to the practical application. Adhering to the proposed theory, we build a plug-and-play module, and the effectiveness of which are substantiated by experiments on benchmarks.
\end{abstract}

\section{Introduction}
Causal inference is a specialized field that presents promising potential with respect to improve machine learning methods, conventionally encompassing four steps: 1) causal model construction for modeling causality in machine learning applications in a qualitative analysis manner~\cite{li2023causal,liu2021learning,chen2022learning}; 2) causal model validation, including independence and causality testing, to demonstrate the correctness of the causal model construction~\cite{daniusis2012inferring,zhang2012kernel,lu2021invariant}; 3) causal model-based deconfounding approach implementation, which prospers in various machine learning fields, e.g., eliminating spurious correlations~\cite{mao2021generative,liu2022show} and performing counterfactual reasoning~\cite{chang2021towards} in computer vision, learning the intrinsic rationale of the graph~\cite{ji2024rethinking,wu2024graph} in graph neural networks, overcoming selection bias~\cite{li2023propensity} and popularity bias~\cite{zhao2022popularity} in the recommendation systems; 4) deconfounding approach estimation improvement, focusing on enhancing the accuracy of causal model-based deconfounding~\cite{zhu2024causal,frauen2023neural}. Benefiting from the establishment of graphical models, the advances of the structural causal model (SCM) concentrate greater potential onto exploring the causal mechanisms behind machine learning techniques, e.g., the analysis of independent relationships among variables and the identification of causal effects for various machine learning applications.

In practice, the involvement of derived discrete data with extra stringent structural constraints increases the complexity of machine learning application scenarios, resulting in a lack of adaptability of conventional interventionism causal analysis theories, i.e., \textit{single-world} SCM-based theory~\cite{xia2021causal,zevcevic2021relating,pawlowski2020deep}. \textit{cross-world} counterfactuals~\cite{richens2022counterfactual,correa2021nested,alomar2023causalsim,shah2022counterfactual} provide a framework to estimate ``what-if'' scenarios that transcend the observed world, aiding in better-informed causal inference, which is crucial for understanding causal relationships in a more comprehensive manner, as it enables the exploration of causality under different hypothetical conditions~\cite{ibeling2020probabilistic,d2022underspecification,shalit2017estimating,khemakhem2021causal,sanchez2021vaca}. A focal issue is the exclusivity of counterfactual variables and real-world variables in an invariant graphical model, challenging the construction of cross-world counterfactual SCMs. In this regard, Twin networks~\cite{graham2019copy,han2022complexity,shpitser2012counterfactuals,vlontzos2023estimating,balke2013counterfactuals} demonstrate a symbiotic relationship of graphical modeling in counterfactual and real-world scenarios. Yet, in this paper, we provide multiple scenarios of cross-world counterfactual causal analysis, where the applications of twin networks \textit{breakdown}, detailed in Section \ref{sec:breakdown}.

To this end, we propose the \textit{teleporter theory} to establish a complete graphical representation of counterfactuals, providing a general and simple approach for modeling cross-world counterfactual causality. Concretely, according to the framework of probabilistic causal models, each variable can ultimately trace its changes back to the exogenous nodes that influence it by iteratively applying the structural equations of its parent nodes over a finite number of steps. Variables that have consistent structural equations in both the real world and the counterfactual world possess equivalence, which is determined as a \textit{teleporter}. Accordingly, we provide sufficient causal analysis from the \textit{structural equation} perspective, substantiating the theoretical correctness of the proposed teleporter theory. In terms of theoretical applications, we focus on two main aspects: 1) we apply $d$-separation to test the conditional independence between any two cross-world variables of a cross-world SCM constructed by introducing the teleporter theory, which can prove the correctness and generalization of our theory; 2) we use the teleporter theory to build the cross-world SCM and further leverage the cross-world symbolic derivation to compute counterfactual probability, which can avoid the complex calculation of the probability distribution of background variables, demonstrating the effectiveness and simplicity of our theory. Adhering to the proposed theory, we build a practical plug-and-play module to address the intrinsic issue in the field of Graph Out-Of-Distribution (GraphOOD) \cite{gui2022good,chen2022learning,jia2024graph}, and the consistency and effectiveness of the proposed module are substantiated by experiments on benchmarks. 

Our \textbf{contributions} are as follows:
(1) We provide multiple motivating examples to elucidate the incompleteness of twin networks in certain cross-world counterfactual scenarios with sufficient causal analysis.
(2) We propose a general and simple approach for modeling cross-world counterfactual causality, namely the teleporter theory, which is proved as a complete causal analysis method.
(3) We provide sufficient evidence to prove the theoretical correctness of the proposed teleporter theory by introducing the structural equation analysis.
(4) We conduct extensive validations on commonly adopted benchmarks, demonstrating the generalized applicability of the teleporter theory from theoretical and practical perspectives.

\section{Related Work}

\paragraph{Modeling Single-World Causality.}
Single-world interventions pertain to the first two levels of Pearl’s causal hierarchy~\cite{pearl2009causality,bareinboim2022pearl}: association and intervention. Once we can model causality using observational data~\cite{jaber2019causal,perkovi2018complete}, various methods exist for estimating interventional distributions~\cite{ke2019learning,xia2021causal,zevcevic2021relating,kocaoglu2017causalgan}, provided identifiability is ensured~\cite{bareinboim2022pearl}. The implementation of interventions transcends simple modeling of data associations, aiming instead to answer scientific questions such as ``How effective is $X$ in influencing $Y$?'' and thus achieving estimates of causal effects. Numerous works in the machine learning community have benefited from this approach: (1) real user preference in recommendation systems, such as deconfounding~\cite{zhang2023debiasing,zhang2023leveraging,he2023addressing} and disentangling~\cite{sun2022multi}, (2) rationale representations in graph neural networks, such as robustness and invariant subgraphs~\cite{wu2022discovering,chen2022learning}, and (3) invariant representations in domain generalization, such as eliminating spurious correlations~\cite{arjovsky2019invariant,cui2022stable}.
\paragraph{Modeling Cross-World Counterfactual Causality.}
Cross-world causality aims to address the top-level query of Pearl’s causal hierarchy~\cite{pearl2009causality,bareinboim2022pearl}: counterfactuals. However, estimating counterfactual causality faces the challenge of conflicts between real-world variable values and counterfactual variable values, making identifiability~\cite{khemakhem2021causal,geffner2022deep,ibeling2020probabilistic,d2022underspecification} more scarce compared to interventions. Despite this, answering counterfactual queries like ``why?'' and ``what if?'' using causal framework can enable personalized and interpretable decision-making and reasoning. This significantly advances several key areas: 1) application in computer vision, e.g., alleviating data scarcity through data augmentation~\cite{kaushik2019learning,xia2022adversarial}; 2) fairness in legal and policy-making contexts~\cite{kusner2017counterfactual,zhang2018fairness}; 3) interpretability in the field of medical health~\cite{richens2022counterfactual,oberst2019counterfactual}, among others.

\section{Preliminary}
\label{p}
We recap the necessary preliminaries of causal background knowledge relevant to our work. For a more in-depth understanding, please refer to the literature~\cite{pearl2009causal,pearl2009causality,pearl2016causal}.

\paragraph{Structural Causal Models.}
A SCM~\cite{pearl2009causality,peters2017elements} is a causal model in a triple form, i.e., $M=\left \langle X,U,F \right \rangle$, where $U$ presents the \textit{exogenous variable set}, determined by external factors of the model. $X=\left \{ X_1,X_2,...,X_n \right \}$ presents the \textit{endogenous variable set}, determined by the internal functions $F=\left \{ f_1,f_2,...,f_n \right \}$. Each $f_i$ represents $\left \{ f_i:U_i\cup PA_i \rightarrow X_i\right \}$, where $U_i\subseteq U$, $PA_i\subseteq X \backslash X_i$, satisfying:
\begin{equation}
    x_i=f_i\left ( pa_i,u_i \right ),\quad i=1,2,...,n.
\end{equation}
$PA_i$ denotes the parent nodes of $X_i$. Note that, in SCM, \textit{uppercase letters} conventionally denote \textit{variables}, and \textit{lowercase letters} conventionally denote \textit{values} of the corresponding variables, e.g., $x_i$ is the value of $X_i$. For ease of discussion, we omit such clarification in the following sections. Each causal model $M$ corresponds to a directed acyclic graph $\mathcal{G}$, where each node corresponds to a variable in $X\cup U$, and directed edges point from $U_i\cup PA_i$ to $X_i$. It is worth noting that exogenous variables $U$ have no ancestor nodes, and each endogenous variable $X_i$ is at least a descendant of one exogenous variable.

\paragraph{Interventions and Do-calculus.}
The causal model $M$ describes intrinsic causal mechanisms, characterized by the observed distribution $P_M(X)=\prod_{i=1}^{n}P(x_i\mid pa_i)$. Intervention\footnote{The definition here refers to the atomic intervention \cite{pearl2009causality}. For brevity, we intervene on only one variable.} is defined as forcing a variable $X_i$ to take on a fixed value $x$, modifying the model $M=\left \langle X,U,F \right \rangle$ to $M_x=\left \langle X,U,F_x \right \rangle$, where $F_x=\left \{ F \backslash f_i  \right \}\cup \left \{ X_i=x \right \}$. This is equivalent to removing $X_i$ from its original functional mechanism $x_i=f_i\left ( pa_i,u_i \right )$ and modifying this function to a constant function $X_i=x$. Formally, we denote the \textit{intervention} as $do(x_i=x)$, called the \textit{$do$-$calculus$}. It explores how causal mechanisms will change when external interventions, or experiments, are introduced. We denote the distribution after the intervention as $P_{M_{x}}(X)=P(x_1,...,x_n\mid do(x_i=x))$, where
\begin{equation}
    P(x_1,...,x_n\mid do(x_i=x))=
    \begin{cases}
    \prod_{j\neq i}P(x_j\mid pa_j)& x_i=x \\
    0& x_i\neq x
    \end{cases}.
\end{equation}

\paragraph{Counterfactuals.}
If $M_x$ defines the effect of the action $do(X=x)$ on $M$, what is the potential change of another endogenous variable $Y$ due to the intervention effect $M_x$? We denote $M_x$ as the SCM of the \textit{counterfactual world}~\cite{pearl2009causality} derived by adopting the intervention $X=x$. The potential value of $Y$ influenced by the intervention $do(X=x)$ is denoted as $Y_x(u)$, which is a solution to the equation set $F_x$, i.e., $Y_x(u)=Y_{M_{x}}(u)$. Concretely, $Y_x(u)$ presents the counterfactual statement ``\textit{Under condition $u$, if $X$ were $x$, then $Y$ would be $Y_x(u)$.}''

\paragraph{Path and $d$-separation.} We recap two classic definitions~\cite{pearl2016causal} to help us determine the independence between variables in the SCM graph. 

\begin{definition}
    \label{path} \textbf{(Path)}
    In the SCM graph, the paths from variable $X$ to $Y$ include three types of structures: 1) chain structure: $A \rightarrow B\rightarrow C$ or $A \leftarrow B\leftarrow C$; 2) fork structure: $A \leftarrow B\rightarrow C$; 3) collider structure: $A \rightarrow B\leftarrow C$.
\end{definition}

\begin{definition}
    \label{separation} \textbf{($d$-separation)}
A path $p$ is blocked by a set of nodes $Z$ if and only if:
\begin{enumerate}
    \item $p$ contains a chain of nodes $A \rightarrow B\rightarrow C$ or a fork  $A \leftarrow B\rightarrow C$ such that the middle node $B$ is in $Z$, i.e., $A$ and $C$ are independent conditional on $B$, or
    \item $p$ contains a collider $A \rightarrow B\leftarrow C$ such that the collider node $B$ is not in $Z$, and no descendant of $B$ is in $Z$, i.e., $A$ and $C$ are independent without conditions.
\end{enumerate}
\end{definition}
If $Z$ blocks every path between two nodes $X$ and $Y$, then $X$ and $Y$ are \textit{$d$-separated}, conditional on $Z$, i.e., $X$ and $Y$ are independent conditional on $Z$, denoted as $X \upmodels Y \mid Z$.

\begin{figure}
    \centering
    \includegraphics[width=\textwidth]{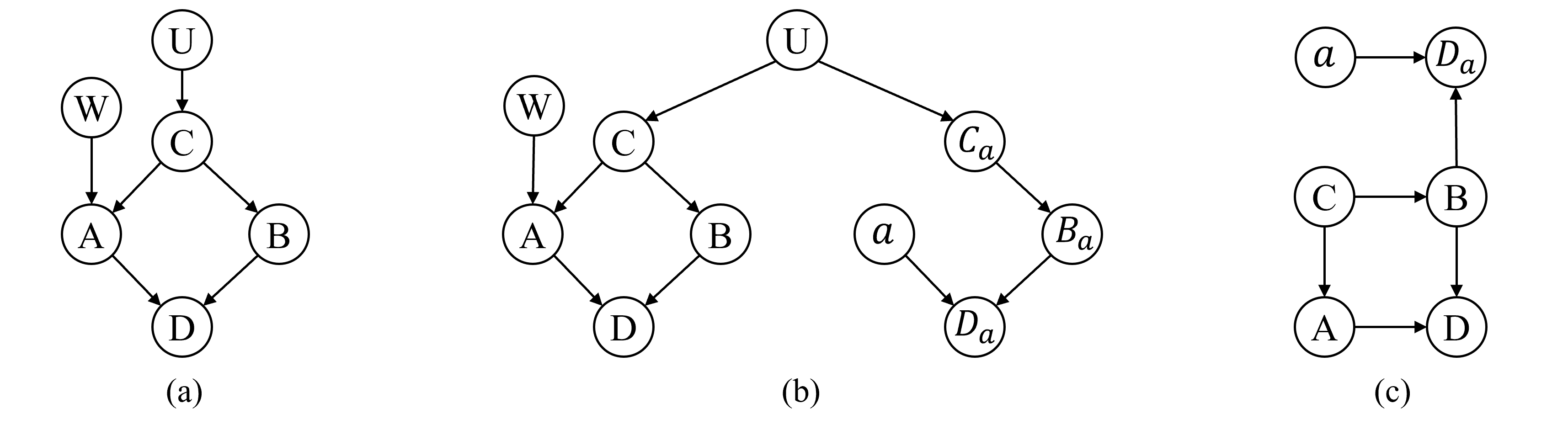}
    \vspace{-0.6cm}
    \caption{Example for breakdown of twin network: Figure (a) represents the real-world SCM, Figure (b) shows the cross-world SCM constructed using the twin network, and Figure (c) illustrates the cross-world SCM constructed using the teleporter theory.}
    \vspace{-0.5cm}
    \label{fig:example2}
\end{figure}

\section{Teleporter Theory}
\label{headings}
In this section, we provide a detailed analysis to show the breakdown of twin networks, and further propose a novel theory in cross-world counterfactual causality modeling, which demonstrates sufficient theoretical completeness and generalization.

\subsection{Theoretical Breakdown of Twin Networks} \label{sec:breakdown}
Twin network~\cite{balke2013counterfactuals} is formed to model cross-world counterfactual causality by connecting the real world and counterfactual world, sharing exogenous variables between them. The constructed sub-networks of real world and counterfactual world are structurally identical, except that the arrows pointing to the intervened variable are removed in the counterfactual sub-network. The specific construction steps are as follows: 1) duplicating the endogenous variables $X = \left \{ X_1,X_2,...,X_n \right \}$ from the real world as endogenous variables $X^{*} = \left \{ X_1^{*},X_2^{*},...,X_n^{*} \right \}$ in the counterfactual world; 2) selecting the intervened variable $X_i$ and removing all arrows pointing to the counterfactual variable $X_i^{*}$; 3) connecting $X$ and $X^{*}$ through exogenous variables $U$ to form the twin network. Figure \ref{fig:example2}(b) illustrates an example of cross-world SCM constructed by using twin networks.

However, the benchmark twin networks encounter \textit{theoretical breakdowns} in certain scenarios of modeling cross-world counterfactual causality. Concretely, we explore an example in ``\textit{Causality}''~\cite{pearl2009causality}, as depicted in Figures \ref{fig:example2}(a) and (b). We aim to test whether $D_a$ is independent of $A$ given $B$ or $C$, i.e., $A \upmodels D_a \mid B$ or $A \upmodels D_a \mid C$. The corresponding twin network of this example causal graph is illustrated in Figure \ref{fig:example2}(b). To assess the conditional independence between $A$ and $D_a$, we determine under which variables $A$ and $D_a$ are $d$-separated. Conditional on $C$, the path from $A$ to $D_a$, i.e., $A \leftarrow C \leftarrow U \rightarrow C_a \rightarrow B_a \rightarrow D_a$, is blocked by node $C$, and thus, $A \upmodels D_a \mid C$ holds. Yet, conditional on $B$, this path is $d$-connected, i.e., $A \nupmodels D_a \mid B$. We validate the conclusions obtained from the twin network using Theorem 4.3.1 from~\cite{pearl2016causal}. Both variables $B$ and $C$ satisfy the back-door criterion for $(A,D)$, indicating that for all values $a$ of $A$, given $B$ or $C$, the counterfactual $D_a$ is conditionally independent of $A$. The above analysis demonstrates that the twin network erroneously determines $A\nupmodels D_a \mid B$, proving that the twin network lacks theoretical completeness in specific cross-world SCMs.

\subsection{Teleporter Theory for Modeling Cross-World Counterfactual Causality} \label{sec:teleporter}
To remedy the mentioned theoretical deficiency, we employ a probabilistic causal model framework~\cite{pearl2009causal} to expound the teleporter theory. As definitions in Section \ref{p}, for SCM $M=\left \langle X,U,F \right \rangle$, a probabilistic causal model is a tuple $\left \langle M,P(u) \right \rangle$, where $P(u)$ is the probability distribution over the set $U$. By the definition of structural equations $x_i=f_i\left ( pa_i,u_i \right )$, the value of an endogenous variable $X_i$ can be recursively represented by all possible values of exogenous variables, i.e., $(u_1,u_2,...,u_n)$, meaning each endogenous variable is a function of $U$. For instance, for a certain endogenous variable $X_i \in X$, we have $P(X_i=x_i)=\sum_{\left \{ u\mid X_i(u)=x_i \right \}}P(u)$. We formalize the above assertion, and the value of $X_i$ can be represented by the following recursively defined function:
\begin{align}
x_i&=f_{X_i}(pa_i,u_i) \\
&=f_{X_i}(f_{X_{i_1}}(pa_{i_1},u_{i_1}),f_{X_{i_2}}(pa_{i_2},u_{i_2}),...,f_{X_{i_k}}(pa_{i_k},u_{i_k}),u_i)\\
&=g_{X_i}(u_1,u_2,...,u_n)
\end{align}
where $pa_i=\left \{ X_{i_{1}},X_{i_{2}},...,X_{i_{k}} \right \}\subset X$, and $g_{X_i}$ is a function determined solely by the exogenous variables $(u_1,u_2,...,u_n)$ after a finite number of iterations.

\begin{figure}
    \centering
    \includegraphics[width=0.92\textwidth]{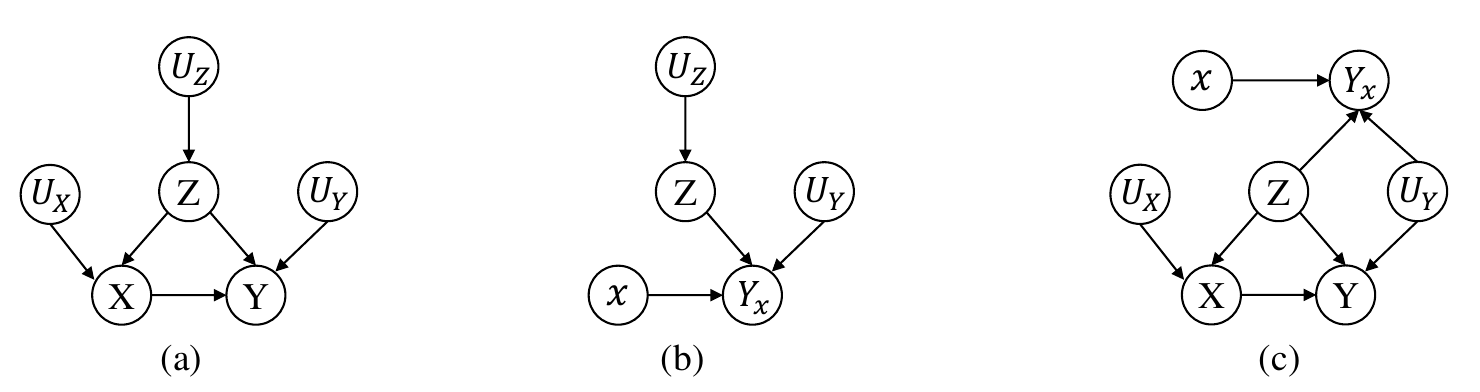}
    \vspace{-0.2cm}
    \caption{Illustration of cross-world SCM construction using teleporter theory: Figure (a) represents the real world $\mathcal{W}_r$, Figure (b) depicts the counterfactual world $\mathcal{W}_c$, and Figure (c) shows the cross-world SCM $\mathcal{W}_m$ formed by connecting the variables in $\mathcal{W}_r$ and $\mathcal{W}_c$ through the teleporter $Z$.}
    \vspace{-0.6cm}
    \label{fig:example1}
\end{figure}

Accordingly, we consider the structural equations of variables in the counterfactual world. Suppose the counterfactual world with the intervention $do(X_i=x^\star)$, and the values of variables $X_j$ are determined as $x_j=f_{X_j}^{x^\star}(pa_j,u_j)=f_{X_j}(f_{X_{j_1}}^{x^\star}(pa_{j_1},u_{j_1}),...,f_{X_{j_k}}^{x^\star}(pa_{j_k},u_{j_k}),u_j)$. The value of $X_i$ is fixed as a constant, i.e., $x_i=f_{X_i}^{x^\star}(pa_i,u_i)=x^\star$. In the counterfactual world, intervening on variable $X_i$ means removing all arrows pointing to $X_i$. Thus, the values of $X_j$ that \textit{only} have $U$ as the parent nodes are $x_j=f_{X_j}^{x^\star}(pa_j,u_j)=f_{X_j}(u_j)$. Only the descendants set $W$ of the intervened variable $X_i$ is influenced by $x_i=x^\star$ according to its structural equation, while the set $Z$ that is $d$-separated from $X_i$ remains the same as in the original world.

For the convenience of counterfactual notation, we present our teleporter theory by introducing \textit{three} endogenous variables as uppercase letters\footnote{Please refer to Section \ref{p} for the definition of counterfactual notation.} $X,Y,...,Z$. Accordingly, we provide the detailed theoretical description of our theory.
\begin{theorem}
    \label{teleporter} \textbf{(Teleporter theory for modeling cross-world counterfactual causality)}
Suppose we intervene on the endogenous variable $X$. Let $\mathcal{W}_r=\left \langle M,u \right \rangle$ denote the real world before the intervention, and $\mathcal{W}_c=\left \langle M_x,u \right \rangle$ denote the counterfactual world, where $u$ denotes the value set of the corresponding exogenous variable set $U$, and $u$ is shared between $\mathcal{W}_r$ and $\mathcal{W}_c$. In the counterfactual world $\mathcal{W}_c$, the variable $Z \subset X\cup U$, that are \textbf{d-separated} from $X=x$, can be determined as the \textbf{teleporter}. $\mathcal{W}_r$ and $\mathcal{W}_c$ are connected through the teleporter $Z$, forming a cross-world SCM $\mathcal{W}_m$. The teleporter theory-based construction of $\mathcal{W}_m$ is required to adhere the following rules:
\begin{itemize}
    \item \textbf{Rule 1}: All descendants of $X=x$ in $\mathcal{W}_r$, e.g., $D$, have the potential value influenced by the intervention $do(X=x)$ in $\mathcal{W}_c$, e.g., $D_x$.
    \item \textbf{Rule 2}: Exogenous variables $U_Z$ associated with teleporter $Z$ should be removed, while the remaining exogenous variables in  $\mathcal{W}_r$ are retained. All teleporters need to be determined in $\mathcal{W}_m$, including exogenous variables in $U$ that are $d$-separated from $X=x$.
\end{itemize}
\end{theorem}
We illustrate the implementation process of teleporter theory by using a classic causal graph as an example. Figure \ref{fig:example1}(a) investigates the causal relationship between $X$ and $Y$, where $Z$ acts as a confounder~\cite{pearl2009causality}, and all exogenous variables are depicted in the graph, which is treated as the real-world SCM $\mathcal{W}_r$. Figure \ref{fig:example1}(b) represents the intervention $do(X=x)$ on $X$, where all arrows pointing to $X$ are removed, which is treated as the counterfactual world $\mathcal{W}_c$. We determine that the structural equation for $Y$ in $\mathcal{W}_c$, denoted as $f_{Y_{x}}(X=x,Z,u_Y)$, is evidently different from the structural equation for $Y$ in $\mathcal{W}_r$, denoted as $f_{Y}(X,Z,u_Y)$. Thus, the meanings and values of $Y$ are different in the two worlds. Therefore, according to \textit{Rule 1} of Theorem~\ref{teleporter}, the descendants of $x$ only consist of $Y$ in $\mathcal{W}_c$, denoted as $Y_x$.

To derive the cross-world SCM, connecting the real-world and counterfactual worlds, we introduce the teleporter theory. According to \textit{Rule 2} of Theorem~\ref{teleporter}, a teleporter refers to a variable with identical structural equations in $\mathcal{W}_r$ and $\mathcal{W}_c$. In $\mathcal{W}_c$, the only endogenous variable, $d$-separated from $X=x$, is $Z$, with its structural equation denoted as $f_{Z}(pa_{Z},u_Z)$. $X$ is \textit{not} the parent node to such variables, and thus they are not influenced by the intervention $do(X=x)$. By utilizing the common teleporter $Z$ shared between $\mathcal{W}_r$ and $\mathcal{W}_c$ as a connecting node, we obtain the cross-world SCM $\mathcal{W}_m$ depicted in Figure \ref{fig:example1}(c).

\section{Theoretical Application of Teleporter Theory}
\label{theory}
The SCM $M$ and its corresponding graph $\mathcal{G}$ facilitate the graphical representation of causal variables, enabling us to intuitively test the independence between variables ($d$-separation). This, in turn, allows us to explore the effects of interventions without conducting new experiments, e.g., back-door/front-door adjustments. However, counterfactual variables $Y_x$ and real-world variables $X$ cannot coexist in a single graph $\mathcal{G}$ due to involving cross-world considerations. Twin network~\cite{balke2022probabilistic,shpitser2012counterfactuals,balke2013counterfactuals} is the first attempt to address this issue, yet such a method fails in certain scenarios, shows its theoretical incompleteness. In the following two subsections, we demonstrate that the teleporter theory can provide a complete graphical representation of counterfactuals.

\subsection{Independence between Cross-World Variables}
The cross-world independence between counterfactual variables and real-world variables is difficult to derive from the separated real-world and counterfactual SCMs or the corresponding structural equations. The significant advantage of the teleporter theory lies in the graphical representation of counterfactuals, enabling us to analyze the (conditional) independence between any pair of cross-world variables. Concretely, considering the breakdown of twin networks in Section \ref{sec:breakdown}, we propose to demonstrate the theoretical completeness and generalization of the proposed teleporter theory as follows. In the cross-world SCM $\mathcal{W}_m$ of Figure \ref{fig:example2}(c) obtained through the teleporter theory, we conclude that both $A\upmodels D_a \mid B$ and $A\upmodels D_a \mid C$ hold. This is because the path from $A$ to $D_a$, i.e., $A \leftarrow C  \rightarrow B \rightarrow D_a$, is blocked by $B$ or $C$. Similarly, upon adding $D$ as a condition, new paths between $A$ and $D_a$ are opened through the collider node $D$. Therefore, conditional on $\left \{ D,C \right \}$, $A$ and $D_a$ are \textit{not} $d$-separated, satisfying $A\nupmodels D_a \mid \left \{ D,C \right \}$. However, conditional on $\left \{ D,B \right \}$, $A$ and $D_a$ are $d$-separated, satisfying $A\upmodels D_a \mid \left \{ D,B \right \}$.

Thus, the proposed teleporter theory can widely empower the (conditional) independence testing between cross-world variables. To better demonstrate the implementation of independence testing in the cross-world SCM built by using the teleporter theory, we propose the following Theorem~\ref{d}, summarizing the d-separation theorem for counterfactuals.
\begin{theorem}
    \label{d} \textbf{($d$-separation for cross-world variables under the teleporter theory)}
In the cross-world SCM $\mathcal{W}_m$ constructed by following the teleporter theory, the path $p$ between the real-world variable $X$ and the counterfactual variable $Y_x$ is $d$-separated by the node set $Z$ if and only if:

    1. $p$ contains either a chain structure or a fork structure, with intermediate nodes in $Z$, or
    
    2. $p$ contains a collider structure, with neither the intermediate node nor its descendants in $Z$.

The set $Z$ $d$-separates $X$ from $Y_x$ if and only if $Z$ blocks all paths from $X$ to $Y_x$.
\end{theorem}

\begin{figure}
    \centering
    \includegraphics[width=0.95\textwidth]{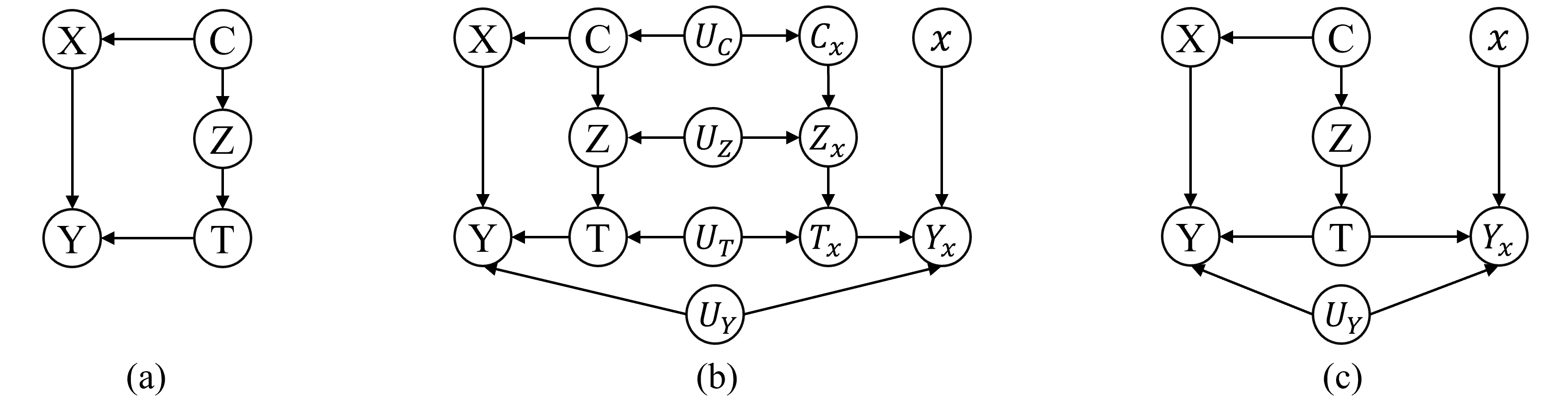}
    \vspace{-0.2cm}
    \caption{Figure (a) represents the real world $\mathcal{W}_r$, Figure (b) shows the cross-world SCM $\mathcal{W}_m$ constructed using the twin network, and Figure (c) depicts the cross-world SCM $\mathcal{W}_m$ constructed using the teleporter theory.}
    \label{fig:example3}
    \vspace{-4ex}
\end{figure}

\subsection{Cross-World Adjustment}
The joint distribution of counterfactual statements requires computation, storage, and utilization of the marginal probability of values of the exogenous variables, i.e., $P(u)$. For example, $P(Y_x=y,X=x')=\sum_{\left \{ u|Y_x(u)=y, X(u)=x' \right \}}^{}P(u)$. Classic works summarize three steps for estimating the counterfactual $P(Y_x\mid e)$ in~\cite{pearl2009causality}, where $e$ denote the observed variable values: 1) abduction: updating $P(u\mid e)$ using the fact $e$; 2) action: updating the SCM $M$ to $M_x$; 3) computing $P(Y_x\mid e)$ in the counterfactual world $\mathcal{W}_c = \left \langle M_x,P(u\mid e) \right \rangle$. However, obtaining the distribution of exogenous variables $U$ is extremely difficult. The teleporter theory provides a simple method to compute $P(Y_x\mid e)$, facilitating cross-world adjustment.

We propose the counterfactual criterion to obtain the (conditional) independence of $X$ and $Y_x$:
\begin{theorem}
    \label{cross} \textbf{(Counterfactual criterion and cross-world adjustment)}
    The evidence $e$ represents the values of the variable $E$ in the real world $\mathcal{W}_r$. Given an observable variable set $Z$, if $E\cup Z$ causes that conditional on $E\cup Z$, $X$ and $Y_x$ are $d$-separated in the cross-world SCM $\mathcal{W}_m$, then the counterfactual $Y_x$ is conditionally independent of $X$, denoted as $X\upmodels Y_x \mid \left \{ E,Z \right \}$. The cross-world adjustment formula can be derived as follows:
\begin{equation}
        P(Y_x=y\mid E=e)=\sum _{z}P(Y=y\mid Z=z,X=x,E=e)P(Z=z \mid E=e).
    \end{equation}
    \end{theorem}
    \vspace{-0.2cm}
In comparison to the counterfactual interpretation of the back-door criterion in~\cite{pearl2016causal}, our theoretical approach is proved to be a generalized solution, since the former approach can only treat the back-door path-related scenarios, our approach can achieve cross-world adjustment for any pair of variables. Please refer to \textbf{Appendix} \ref{proof} for the corresponding proofs.

We will now present two examples to illustrate that the teleporter theory is more complete compared to twin networks, as the latter fails in multiple scenarios. The first example demonstrates cases where twin networks \textit{incorrectly} identify the required variables for adjustment. The SCM of such an example is depicted in Figure \ref{fig:example3}(a). In the twin network of Figure \ref{fig:example3}(b), $X$ and $Y_x$ are $d$-connected by the path $X\leftarrow C \leftarrow U_c \rightarrow C_x \rightarrow Z_x \rightarrow T_x \rightarrow Y_x$. If we acquire to compute $Y_x$, the variables for adjustment can only be $C$, since using $Z$ or $T$ for adjustment would open up a collider node, resulting in certain dependence relationships between the parent nodes. However, in the cross-world SCM $\mathcal{W}_m$ of Figure \ref{fig:example3}(c) modeled by using the teleporter theory, the path between $X$ and $Y_x$ is $X\leftarrow C \rightarrow Z \rightarrow T \rightarrow Y_x$. According to Theorem~\ref{cross}, we can perform the adjustment on any node in $\left \{ C,Z,T \right \}$, which is consistent with the empirical conclusion in ``\textit{Causality}''~\cite{pearl2009causality}.

\begin{figure}
    \centering
    \includegraphics[width=0.95\textwidth]{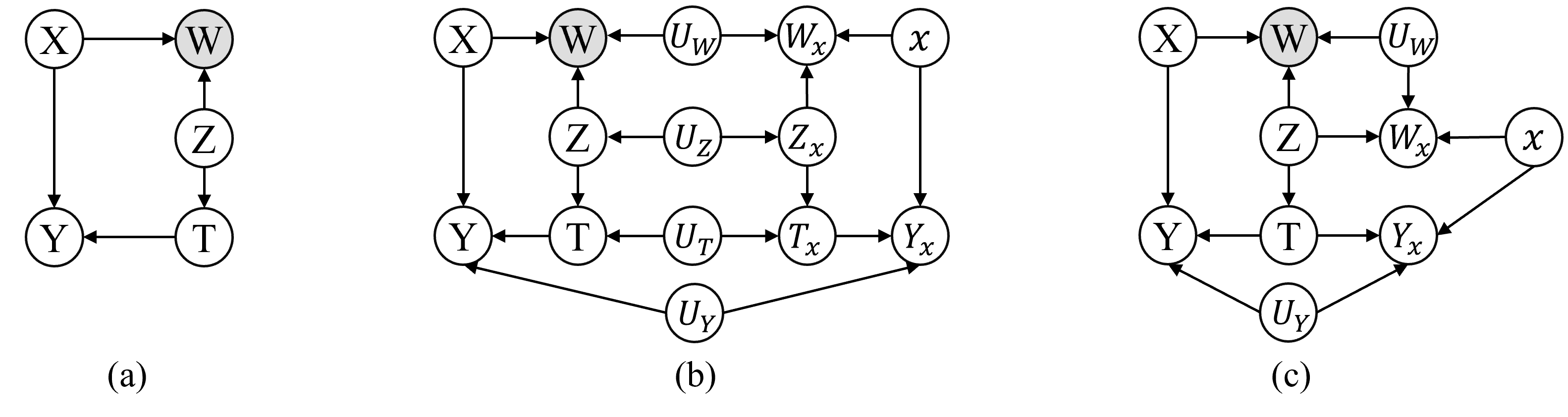}
    \vspace{-0.3cm}
    \caption{Figure (a) represents the real world $\mathcal{W}_r$, Figure (b) shows the cross-world SCM $\mathcal{W}_m$ constructed using the twin network, and Figure (c) depicts the cross-world SCM $\mathcal{W}_m$ constructed using the teleporter theory. Grey nodes indicate conditioning on that variable}
    \vspace{-3ex}
    \label{fig:example4}
\end{figure}

The second example, as illustrated in Figure \ref{fig:example4}(a), involves computing $P(Y_x\mid w)$ given the known evidence $w$. In the twin network of Figure \ref{fig:example4}(b), $X$ and $Y_x$ are connected only through one path: $X \rightarrow \underline{W} \leftarrow Z \leftarrow U_z \rightarrow Z_x \rightarrow T_x \rightarrow Y_x$\footnote{\underline{W} represents conditioning on W, i.e., W is given.}. In this case, we can only perform the adjustment on $Z$, since the adjustment on $T$ would open up new paths, i.e., $Z\nupmodels U_T \mid T$. The cross-world SCM $\mathcal{W}_m$ constructed by using the teleporter theory is depicted in Figure \ref{fig:example4}(c), and according to Theorem~\ref{cross}, we can perform the adjustment on both $T$ and $Z$, which well fits the empirical conclusion in ``\textit{Causality}''~\cite{pearl2009causality}. The above theoretical application analysis sufficiently demonstrate the generalization and applicability of our teleporter theory.

\section{Practical Application of Teleporter Theory}
\label{method}
To demonstrate the practical applicability of teleporter theory, we utilize our theory to model the set of variables in GraphOOD \cite{gui2022good,chen2022learning,jia2024graph}, and propose a method to perform the counterfactual conditional probability estimation between cross-world variables. 


\textbf{Preliminary of GraphOOD}. The area of OOD learning deals with scenarios in which training and test data follow different distributions. Furthermore, GraphOOD problems focus on not only general feature distribution shifts but also structure distribution shifts \cite{gui2022good}. Accordingly, Graph neural networks (GNNs) \cite{kipf2016semi,xu2018powerful} are designed based on node features and adjacent matrix to pass messages, which perform well in solving GraphOOD problems. Previous GNN-based methods \cite{yang2022learning,chen2022learning,zhuang2024learning} aim at disentangling the invariant part and the environment part of the input graph to find rationales, thereby addressing the problem of domain shift. The common learning paradigm is shown in Figure \ref{fig:architecture}, where the input graph is processed and split into two latent variables, i.e., the invariant representation $z^{Inv}$ and the environment-dependent representation $z^{Env}$, and only $z^{Inv}$ is used for label prediction, which is achieved by leveraging the empirical observation encompassing the available labeled samples.

\subsection{Cross-World Counterfactual Causality Modeling via Teleporter Theory}
The obtained invariant representation unavoidably contains significant environment-dependent information due to the inherent \textit{inductive bias} arisen from the learning paradigm of benchmark methods. On the contrary, the desired invariant representation is expected to solely contain pure environment-agnostic predictive information. However, such a representation can barely be acquired in the real world yet feasibly obtained in the counterfactual world.

\begin{figure}
    \centering
    \includegraphics[width=0.85\textwidth]
    {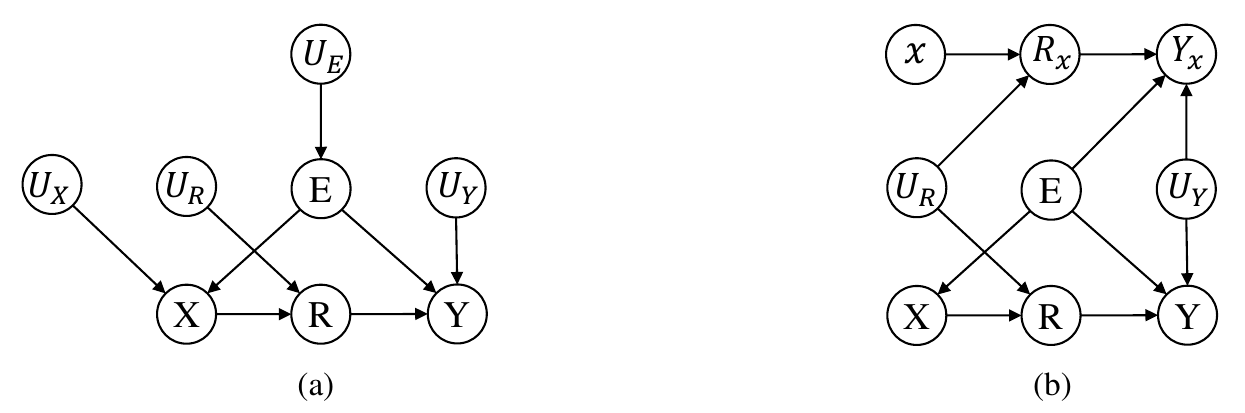}
    \vspace{-0.2cm}
    \caption{SCM for GraphOOD. Figure (a) denotes the real-world SCM. Figure (b) denotes the cross-world SCM.}
\vspace{-0.6cm}
    \label{fig:SCM4GraphOOD}
\end{figure}

To this end, we propose to explore the causal mechanism behind both real-world and counterfactual variables, which is accomplished by modeling the cross-world counterfactual causality. Concretely, we establish an SCM at first, as depicted in Figure \ref{fig:SCM4GraphOOD}(a), which elaborates on the causality among the variables in GraphOOD in the real world. In Figure \ref{fig:SCM4GraphOOD}(a), there exist four endogenous variables in the real world: the input graph $X$, the learned representation $R$ of $X$, the predicted label $Y$ and the environment-dependent information $E$. $U_X$, $U_R$, $U_E$ and $U_Y$ are four exogenous variables corresponding to the endogenous variables. According to Theorem \ref{teleporter}, the variable $E$ can be determined as the teleporter, so the cross-world SCM is demonstrated in Figure \ref{fig:SCM4GraphOOD}(b), where $x$ represents the intrinsic causal subgraph, $R_{x}$ denotes the environment-agnostic invariant representation, and $Y_{x}$ denotes the predicted label corresponding to the graph $x$, which is also the \textit{true} label, since ideally, $x$ and $R_x$ only include environment-agnostic task-dependent information in the counterfactual world.


\subsection{Counterfactual Conditional Probability Estimation via Multi-Scale Mixup Scheme}
\begin{figure}
    \centering
    \includegraphics[width=\textwidth]{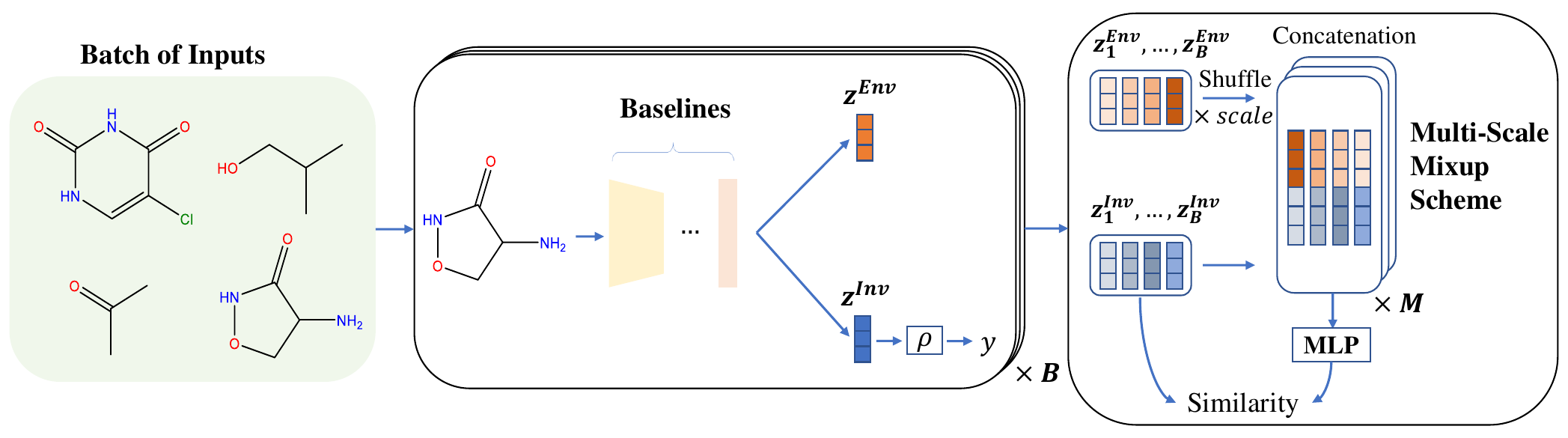}
    \vspace{-0.5cm}
    \caption{GraphOOD learning paradigm with Multi-Scale Environment Mixup Scheme.}
    \vspace{-0.6cm}
    \label{fig:architecture}
\end{figure}

According to the cross-world SCM in Figure \ref{fig:SCM4GraphOOD}(b), we determine that our objective is to derive the \textit{ideal} predicted label $Y_x$ by only using the \textit{available} $X$. Such an objective can be formalized as follows: computing the counterfactual probability $P(Y_x = y \mid X = x')$, where $x'$ denotes the available value of $X$, and $y$ denotes the true label of $x'$. Deriving $P(Y_x = y \mid X = x')$ is equivalent to calculating the conditional probability of $X$ on $Y_x$ in the cross-world SCM. Such a computation can be approximated by using neural network-based methods. Adhering Theorem \ref{d}'s $d$-separation for cross-world counterfactuals, we can directly obtain $X \upmodels Y_{x} \mid E$ from the cross-world SCM in Figure \ref{fig:SCM4GraphOOD}(b). The calculation of $P(Y_{x} = y \mid X = x')$ can be acquired as follows:
\begin{equation}
P(Y_x=y\mid X=x')=\sum_e P(Y_x=y\mid X=x,E=e)P(E=e\mid X=x') \label{eq:ccp}
\end{equation}
As demonstrated in Figure \ref{fig:architecture}, to acquire $P(Y_{x} = y \mid X = x')$, we propose to design a fine-grained method, which can derive the invariant part $z^{Inv}$ and the environment-dependent part $Z^{Env}$ from the input graph $x'$, thereby predict the true label $y$ by leveraging $Z^{Inv}$. According to Equation \ref{eq:ccp}, $P(Y_{x} = y \mid X = x')$ can be estimated by summing the conditional probability of $P(Y_x=y\mid X=x,E=e)$ with respect to different environment-dependent information $E$, i.e., $z^{Env}$. Following \cite{zhuang2024learning}, we utilize a contrastive learning module to estimate $P(Y_x=y\mid X=x,E=e)$, where $z^{Inv}$ is firstly concatenated by a shuffled batch of $z^{Env}$, and then projected into $\tilde{z}^{Inv}$ via a MLP-predictor $\rho$. Ultimately, $z^{Inv}$ and $\tilde{z}^{Inv}$ are used to measure the similarity for contrasting. Accordingly, expanding the available value set of $E$ can widely obtain a more precise estimation of $P(Y_{x} = y \mid X = x')$ in Equation \ref{eq:ccp}. Hence, we introduce a Multi-Scale Mixup Scheme (MsMs) to enrich the available data of $E$, which is achieved by leveraging a hyperparameter $scale$ in concatenating the shuffled environment $z^{Env}$. Furthermore, we further expand the available value set of $E$ by the scaled mixup scheme by $M$ times.

\subsection{Experiments on GraphOOD}
The detailed descriptions of the benchmarks and the baselines are in \textbf{Appendix} \ref{app:benchmarks} and \textbf{Appendix} \ref{app:baselines}, respectively.
To ensure reproducibility, the intricate details of our method’s architecture, and our hyper-parameter settings are detailed in the \textbf{Appendix} \ref{app:hyper-parameters}. The empirical results on the GOOD and DrugOOD benchmarks are presented in Table \ref{tab:good}. By enhancing the available value set $E$ with MsMs, our method places the best in four of six datasets, and shows the best average ROC-AUC score among the baselines, which indicates the effectiveness of our proposed method and further emphasizes the practical generalization of teleporter theory.

\begin{table}[t]
\centering
\small
\setlength\tabcolsep{3pt}
\caption{Evaluation performance on GOOD \cite{gui2022good} and DrugOOD \cite{ji2022drugood} benchmark. The best is marked with \textbf{boldface} and the second best is with \underline{underline}. $\dag$ denotes the reproduction results.}
\scalebox{0.9}{{
\begin{tabular}{l|cc|cccc|c}
\toprule
\multirow{2}{*}{Method} & \multicolumn{2}{c|}{GOOD-HIV} & \multicolumn{4}{c|}{DrugOOD} & \multirow{2}{*}{Average} \\ 
                        & scaffold-covariate & size-covariate & IC50-assay & IC50-scaffold & EC50-assay & EC50-scaffold \\ \midrule
DIR \cite{wu2022discovering}                    & 68.44(2.51)   & 57.67(3.75) & 69.84(1.41)                 & 66.33(0.65)             & 65.81(2.93)               &63.76(3.22)  & 65.31                 \\
GSAT \cite{miao2022interpretable}                   & 70.07(1.76)   & 60.73(2.39) & 70.59(0.43)               & 66.45(0.50)                  & 73.82(2.62)               &64.25(0.63)   &   67.65               \\
GREA \cite{liu2022graph}                   & 71.98(2.87)& 60.11(1.07) & 70.23(1.17)               & 67.02(0.28)                  & 74.17(1.47)               &64.50(0.78)       &  68.00             \\
CAL \cite{sui2022causal}                    & 69.12(1.10)   & 59.34(2.14) & 70.09(1.03)               & 65.90(1.04)                  & 74.54(4.18)&65.19(0.87)     & 67.36            \\
DisC \cite{fan2022debiasing}                   & 58.85(7.26)   & 49.33(3.84) & 61.40(2.56)               & 62.70(2.11)                  & 63.71(5.56)               &60.57(2.27)   & 59.42               \\
MoleOOD \cite{yang2022learning}                & 69.39(3.43)   & 58.63(1.78) & 71.62(0.52)               & \underline{68.58(1.14)}& 72.69(1.46)               &65.74(1.47)     &  67.78             \\
CIGA \cite{chen2022learning}                   & 69.40(1.97)   & 61.81(1.68) & \textbf{71.86(1.37)}& \textbf{69.14(0.70)}                  & 69.15(5.79)               &\underline{67.32(1.35)}    &   68.11             \\
 iMoLD \dag \cite{zhuang2024learning} & \underline{73.54(1.33)}& \underline{65.87(1.98)}& 71.23(0.14)& 67.30(0.35) & \underline{76.03(1.66)}&66.41(1.88) &  \underline{70.06}\\
iMoLD+MsMs & \textbf{74.43(1.96)}& \textbf{66.19(2.32)}& \underline{71.70(0.62)}&  67.77(0.48)& \textbf{77.29(0.65)}&\textbf{67.79(0.84)} & \textbf{70.86}\\ \bottomrule
\end{tabular}
}}
\vspace{-0.7cm}
\label{tab:good}
\end{table}

\section{Conclusions and Limitations}
\label{con}
We strive to explore graphical representation of counterfactuals and propose the teleporter theory to address the challenge of simultaneously representing real-world and counterfactual variables in a single SCM. The cross-world SCM constructed by using the teleporter nodes can well avoid the theoretical breakdown of twin networks in various cross-world counterfactual scenarios, thereby demonstrating the completeness and generalization of the teleporter theory. 
However, the rules that teleporter variables are required to adhere are quite stringent, and introducing such constraints increases the complexity of constructing cross-world SCMs. In future work, we will explore to simplify the proposed rules for determining teleporter variables and attempt to apply our theory in the scenarios involving multiple counterfactual worlds.


\bibliography{ref}
\bibliographystyle{unsrt}
\appendix

\newpage
\section{Proof of Theorem 3}
\label{proof}
Below, we use the calculation of the counterfactual statement $P(Y_x=y \mid E=e)$ as an example to illustrate that once the conditional independence of relevant variables is obtained through cross-world SCM, cross-world adjustment can be achieved using simple algebraic derivations:
\begin{align}
P(Y_x=y\mid E=e)&=\sum_z P(Y_x=y\mid E=e,Z=z)P(Z=z\mid E=e) \\
&=\sum_z P(Y_x=y\mid X=x,E=e,Z=z)P(Z=z\mid E=e) \label{a1} \\
&=\sum_z P(Y=y\mid X=x,E=e,Z=z)P(Z=z\mid E=e). \label{a2}
\end{align}

Equation \ref{a1} holds because $X \upmodels Y_x \mid \{E, Z\}$. Equation \ref{a2} holds due to the consistency condition: $X(u)=x, Y(u)=y \rightarrow Y_x(u)=y$.

\section{Experimental Settings}
\subsection{Benchmarks}\label{app:benchmarks}

We employ two real-world GraphOOD benchmarks, i.e. GOOD \cite{gui2022good} and DrugOOD \cite{ji2022drugood} to exam the performance of our method. GOOD is a systematic benchmark which is tailored specifically for graph OOD problems. We adopt one molecular dataset GOOD-HIV for the graph prediction task, where the objective is binary classification to predict whether a molecule can inhibit HIV.
DrugOOD is an OOD benchmark for AI-aided drug discovery, which provides 
two measurements (IC50 and EC50) and their environment-splitting strategies (assay, scaffold, and size).  According to the split strategy, we choose four datasets, e.g., IC50-assay, IC50-scaffold, EC50-assay, EC50-scaffold as the benchmarks. Due to the chosen task of GOOD-HIV and DrugOOD are both binary classification, we adopt the ROC-AUC score as the evaluation metric.
The details of benchmark are shown in Table \ref{tab:benchmarks}.

\begin{table}[h]
\caption{Benchmark statistics. BC denotes Binary Classification.}
\small
\centering
\begin{tabular}{cccc|cccccc}
\toprule
\multicolumn{3}{c|}{Dataset}                                                                             & Task                             & Metric    & \#Train   & \#Val     & \#Test    & \#Tasks \\ \midrule
\multicolumn{1}{c|}{\multirow{2}{*}{GOOD}}   
& \multirow{2}{*}{HIV}  & \multicolumn{1}{c|}{scaffold-covariate} & BC  & ROC-AUC & 24682   & 4133  & 4108   & 1   \\
\multicolumn{1}{c|}{} 
& & \multicolumn{1}{c|}{size-covariate} & BC  & ROC-AUC & 26169   & 4112  & 3961   & 1    \\
\midrule
\multicolumn{1}{c|}{\multirow{4}{*}{DrugOOD}} & \multirow{2}{*}{IC50} & \multicolumn{1}{c|}{assay}            & BC            & ROC-AUC & 34953   & 19475 & 19463  & 1       \\
\multicolumn{1}{c|}{}   &    & \multicolumn{1}{c|}{scaffold} & BC & ROC-AUC & 22025   & 19478 & 19480  & 1       \\
\cmidrule(){2-9} 
\multicolumn{1}{c|}{} 
& \multirow{2}{*}{EC50} & \multicolumn{1}{c|}{assay}  & BC & ROC-AUC & 4978 & 2761 & 2725 & 1 \\
\multicolumn{1}{c|}{} 
& & \multicolumn{1}{c|}{scaffold} & BC & ROC-AUC & 2743 & 2723  & 2762   & 1       \\                    \bottomrule
\end{tabular}
\label{tab:benchmarks}

\end{table}

\subsection{Baselines}\label{app:baselines}

To compare our method with other methods, we include three interpretable graph learning methods (DIR \cite{wu2022discovering}, GSAT \cite{miao2022interpretable} and GREA \cite{liu2022graph}) and five GraphOOD algorithms (CAL \cite{sui2022causal}, DisC \cite{fan2022debiasing}, MoleOOD \cite{yang2022learning}, CIGA \cite{chen2022learning} and iMoLD \cite{zhuang2024learning}) as baselines.
Note that iMoLD performs not stable on DrugOOD benchmarks, so we reproduce the results using the official code on github.
The descriptions and the github links of the baselines are listed as follows:
\begin{itemize}[leftmargin=*]
\item \textbf{DIR}~\cite{wu2022discovering} identifies an invariant rationale by performing interventional data augmentation to generate multiple distributions from the subset of a graph.
\url{https://github.com/Wuyxin/DIR-GNN}
\item \textbf{GSAT}~\cite{miao2022interpretable} introduces an interpretable graph learning method that leverages the attention mechanism. It injects stochasticity into the attention process to select subgraphs relevant to the target labels. \url{https://github.com/Graph-COM/GSAT}
\item \textbf{GREA}~\cite{liu2022graph} identifies subgraph structures called rationales by employing an environment replacement technique. This allows the generation of virtual data points, which in turn enhances the model's generalizability and interpretability. \url{https://github.com/liugangcode/GREA}
\item \textbf{CAL}~\cite{sui2022causal} introduces a causal attention learning strategy for graph classification tasks. This approach encourages GNNs to focus on causal features, while mitigating the impact of shortcut paths. \url{https://github.com/yongduosui/CAL}
\item \textbf{DisC}~\cite{fan2022debiasing} takes a causal perspective to analyze the generalization problem of GNNs. It proposes a disentangling framework that learns to separate causal substructures from biased substructures within graph data. \url{https://github.com/googlebaba/DisC}
\item \textbf{MoleOOD}~\cite{yang2022learning} investigates the OOD problem in the domain of molecules. It designs an environment inference model and a substructure attention model to learn environment-invariant molecular substructures. \url{https://github.com/yangnianzu0515/MoleOOD}
\item \textbf{CIGA}~\cite{chen2022learning} proposes an information-theoretic objective that extracts the desired invariant subgraphs from the causal perspective. \url{https://github.com/LFhase/CIGA}
\item \textbf{iMoLD} ~\cite{zhuang2024learning} propose a first-encoding-then-split method to disentangle the invariant representation and the environment representation via a residual vector quantization skill and a self-supervised learning pattern. \url{https://github.com/HICAI-ZJU/iMoLD} 
\end{itemize}

\subsection{Hyper-Parameters}\label{app:hyper-parameters}
We reproduce iMoLD with the best hyper-parameters provided in the paper. As for the MsMs part, we choose $scale$ from \{0.3, 0.7, 1.0\}, $M$ from \{1, 3, 5\}.

\subsection{Experiments Compute Resources}
\label{Resources}
Experiments are conducted on one 24GB NVIDIA RTX 4090 GPU.

\end{document}